\documentclass[10pt,twocolumn,letterpaper]{article}

\usepackage[pagebackref=true,breaklinks=true,letterpaper=true,colorlinks,bookmarks=false]{hyperref}
\usepackage{cvpr}
\usepackage{times}
\usepackage{epsfig}
\usepackage{graphicx}
\usepackage{amsmath}
\usepackage{amssymb}
\usepackage{blindtext}
\usepackage{graphicx}
\usepackage{caption}
\usepackage{subcaption}
\usepackage{enumitem}
\usepackage{array}
\usepackage{placeins}
\usepackage{multirow}
\usepackage{booktabs}
\usepackage{float}
\usepackage[table,xcdraw]{xcolor}

\usepackage{cite}
\usepackage{blindtext}
\usepackage{abstract}

\cvprfinalcopy 

\def\cvprPaperID{****} 

\begin{document}

\title{Nested Scale-Editing for Conditional Image Synthesis}


\newcommand \CoAuthorMark{\footnotemark[\arabic{footnote}]} 

\author{Lingzhi Zhang\thanks{equal contribution.} \qquad Jiancong Wang\protect\CoAuthorMark \qquad Yinshuang Xu\qquad Jie Min\qquad \\ Tarmily Wen\qquad James C. Gee\qquad Jianbo Shi \\ \\University of Pennsylvania\\}








\twocolumn[{%
\renewcommand\twocolumn[1][]{#1}%
\maketitle
\begin{center}\vspace{-0.9cm}
    \centering
    \includegraphics[width=\textwidth]{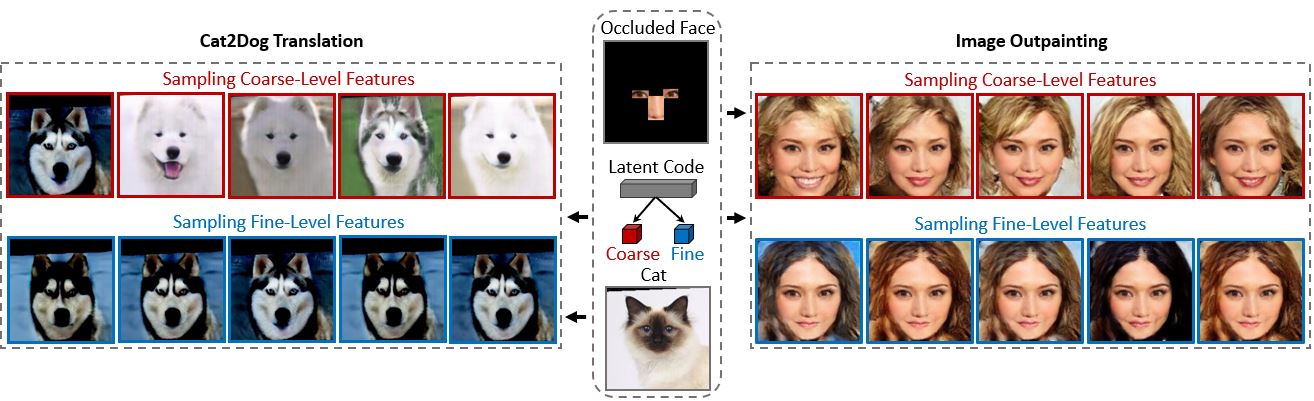}
    \captionof{figure}{ Our approach enables scale-specific visual editings in conditional image synthesis.  We can choose to surgically manipulate coarse-level structural information or fine-level details in Cat2Dog translation and image outpainting tasks.}
    \label{fig:header}
\end{center}%
}]

\saythanks

\begin{abstract}

\noindent
We propose an image synthesis approach that provides stratified navigation in the latent code space. With a tiny amount of partial or very low-resolution image, our approach can consistently out-perform state-of-the-art counterparts in terms of generating the closest sampled image to the ground truth. We achieve this through scale-independent editing while expanding scale-specific diversity. Scale-independence is achieved with a nested scale disentanglement loss. Scale-specific diversity is created by incorporating a progressive diversification constraint. We introduce semantic persistency across the scales by sharing common latent codes. Together they provide better control of the image synthesis process. We evaluate the effectiveness of our proposed approach through various tasks, including image outpainting, image superresolution, and cross-domain image translation.

\end{abstract}

\vspace{-10 pt}
\section{Introduction}

\noindent
Imagine that we want to identify a person based on the appearance of their eyes and nose, or a lower resolution image, as shown in figure \ref{fig:header}.  One solution may be to outpaint the entire face, conditioned on the partial information available. We want to be as imaginative and as detailed as possible to give us a greater chance of success in finding the right person. These tasks are multimodal in nature, i.e., a single input corresponds to many plausible outputs.

Conditional image synthesis approaches aim to solve this problem by sampling stochastic latent codes to generate images in a GAN setting. However, these image synthesis methods of sampling operate as uncontrollable “black boxes”. During inference, we can only hope that a sampled random variable generates the ideal image we desire; otherwise, we need to keep sampling.




\begin{figure*}
\centering
\includegraphics[width=\textwidth]{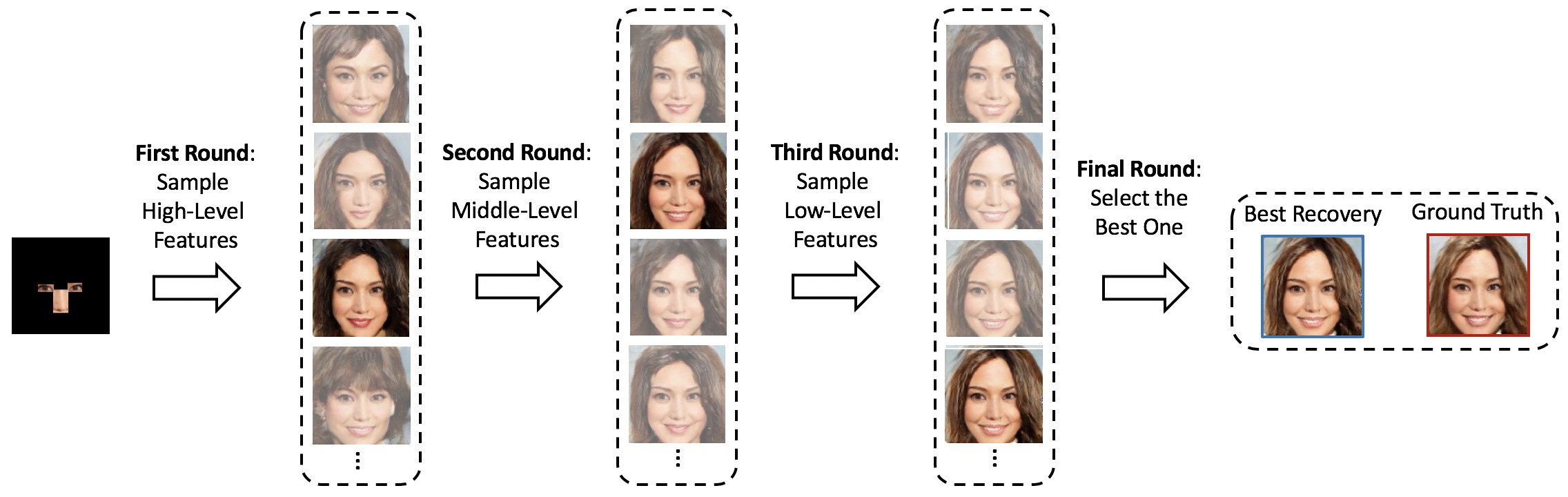}
\caption{Demonstration of an application scenario in which a user interactively recovers a facial identity by sampling scale-specific visual details using our proposed approach. Image you look at this occluded face, you might have a rough mental picture of someone. We can edit the image at multi-scale to recover the identity. }
\label{fig:recovery_app}
\vspace{-10 pt}
\end{figure*}

We propose a steerable conditional image synthesis approach. Inspired by the steerable filtering in the wavelet process \cite{simoncelli1995steerable}, we wish to `steer' the image synthesis across the spatial scales consistently. While in steerable filtering we are concerned with angular edge orientation, in our domain, we focus on object semantics. Specifically, we aim to create visual information from a coarse-level structure to fine-level texture. The key objectives are 1) {\bf scale-independence}: we learn disentangled representations that model scale-specific visual details, and 2) {\bf diversity/mode covering}: we ensure that the decoder covers diverse variations presented on ground truth images.



To implement the scale-independent objective, we take inspiration from the Laplacian image pyramid decomposition: our algorithm essentially learns to generate progressively more refined image along  spatial scales, with each level of refinement independent of each other. To implement the diversity objective, we extend a successful diversity constraint \cite{ndiv} to multi-scale and ensure scale-specific diversity.

Unlike current multi-scale noise injection methods \cite{style-based, denton2015deep}, our multi-scale injected noises share {\em same} latent variable during training. This introduces semantic persistency, meaning the decoder expects latent variables on different scales to have similar semantic meaning. Semantic-persistence can play a major role in search efficiency, since it enables stratified navigation in the latent code space. Fig.\ref{fig:recovery_app} illustrates the stratified navigation process for face superresolution. We first coarsely sample a widespread set of latent random variables to find an image roughly matching the ground truth. Because of our scale-independent representation, we can efficiently edit the image by adjusting any of the latent variable at a specific scale and edit the image information at the corresponding scale. Therefore we can generate a refined image by adjusting the existing latent variable at next scale and repeat, until final scale is reached. This is the ideal steering behavior we seek.

In summary, we highlight our contributions as follows:
\begin{itemize}[noitemsep]
  \item We are the first to propose a multi-scale feature disentanglement loss and a progressive diversification regularization to achieve scale-specific control for conditional image synthesis. 
  \item To the best of our knowledge, our work is the first to utilize diverse conditional image generation for identity recovery. We developed three evaluation metrics for identity recovery in diverse conditional synthesis scenarios. 
  \item We evaluate our aforementioned development on tasks of image outpainting, image superresolution, and multimodal image translation. Our methods achieves competitive image quality and diversity compared to state-of-the-art counterparts, while consistently outperforming them in terms of identity recovery. 
\end{itemize}

\section{Related Work}

\subsection{Multimodal Conditional Generation}

\noindent
Deep generative models have been widely used in many conditional image synthesis tasks, such as super-resolution \cite{SR1, SR2, SR4, SR5, SR6, sr7}, inpainting missing regions \cite{inpaint2, inpaint3, inpaint4, inpaint5, inpaint6, inpaint7, inpaint8, zhang2020multimodal}, style transfer \cite{style1, SR3_style2, style3, style4, style5}, image blending\cite{zhang2020deep, wu2019gp, luan2018deep, tsai2017deep}, and text-to-image \cite{text2img_1, text2img_2, text2img_3}. The majority of these tasks are in nature multimodal, where single input condition may correspond to multiple plausible outputs. BicycleGAN \cite{bicyclegan} first proposed to model this one-to-many distribution by explicitly encoding the target domain into a compact Gaussian latent distribution from which the generator samples. During inference, the generator maps a random variable drawn from the latent distribution, combined with the given input, to the output. StackedGAN and its variant StackedGAN++ \cite{zhang2017stackgan, zhang2018stackgan++} proposed to use a hierarchical generator which incorporates conditional code on multiple scales and were able to generate high quality synthetic images. DRIT \cite{drit} and MUNIT \cite{muint} proposed to disentangle the features into domain-invariant content codes and domain-specific style codes for unpaired image-to-image translations. During inference, the sampled style codes combined with the content code can be transformed into many plausible outputs.

Although the above approaches can generate multimodal outputs given a conditional input, there is no explicit constraint to prevent the generator from mapping the various sampled random variables to similar outputs, which is known as mode collapse. Two concurrent works aim to alleviate this issue by proposing diversity regularization techniques for generative training. Mode Seeking GAN (MSGAN) \cite{msgan} proposes to maximize the ratio of two sampled images over the corresponding latent variables. Normalized Diversification (NDiv) \cite{ndiv} proposes to enforce the generator to preserve the normalized pairwise distance between the sparse samples from a latent distribution to the corresponding high-dimensional output space.

In addition to image synthesis, other applications with multimodal predictions include but are not limited to predicting uncertain motion flows in the future \cite{future_motion_1, future_motion_2, future_motion_3}, hallucinating diverse body pose affordance in 2D \cite{human_affordance_1} and 3D \cite{human_affordance_2} scenes. 

\begin{figure*}
\centering
\includegraphics[width=\textwidth]{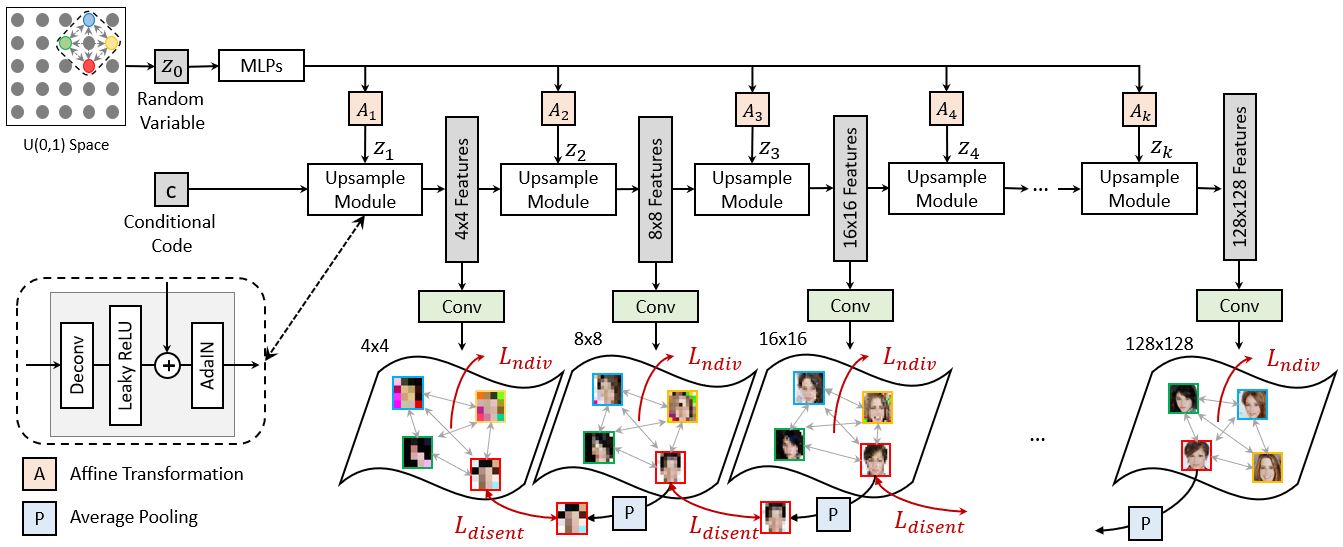}
\caption{Our decoder network takes a conditional code and a single random variable as inputs. A single latent variable is injected into multi-scale feature representations through the first several shared layers of MLPs. Then that output is injected into different affine transformation layers with Adaptive Instance Normalization (AdaIN) \cite{huang2017arbitrary}. At every spatial scale, a single convolutional layer is used to decode a real image at each corresponding resolution. Our proposed disentanglement loss enforces the generated images at every scale to be averaged-pooled back into a lower-resolution generated image. At each iteration, we sample four latent variables and generate four images at each spatial scale, where we also enforce the pairwise distance between sampled images and latent variables to encourage diverse synthesized outputs. }
\label{fig:main}
\end{figure*}

Different from previous work, our focus is to achieve scale-specific control for image synthesis and progressively inject stochasticity into different scales of the synthesized image. Therefore, our proposed techniques can be readily added into these orthogonal previous approaches. We demonstrate that our techniques work with face out-painting, face superresolution, and multimodal animal translation with modified MUNIT \cite{muint}.

\subsection{Feature Disentanglement}

\noindent
Recently there is increasing interest in disentanglement of distinct image characteristics for image synthesis. \cite{disconsty1, disconsty2,disconsty3} attempt to decouple image style and content, while \cite{disshap1,disshap2,disshap3} target object shape and appearance. 
These approaches explicitly incorporate two codes that denote the two characteristics, respectively, into the generative model and introduce a guided loss or incorporate the invariance constraint to orthogonalize the two codes, while our method disentangles the scale-specific variations from global structural feature to local texture feature through hierarchical input of latent variables into the generative model. 

Besides the disentanglement of two specific characteristics, several prior efforts \cite {infogan,betaVAE,metric3, 2016adv,2016adv2} have explored partially or fully interpretable representations of independent factors for generative modeling. Some current work learn representations of the specific attribute by supervised learning \cite{supdis1,supdis2} with a conditioning discriminator. Our method focuses on multi-scale disentanglement through unsupervised learning, which is distinct from the concept of disentangling specific semantic factors. 

\cite{karras2017progressive} has first proposed synthesize images from low-resolution scale to higher-resolution scale in a progressive manner. Similarly, other closely related work is \cite{denton2015deep} and \cite{style-based}. \cite{denton2015deep} pioneered the use of the hierarchical generator with latent codes injected at each level for multi-scale control of image synthesis and further advanced in \cite{style-based}. Both \cite{denton2015deep} and \cite{style-based}, however, only target unconditional image synthesis and do not explicitly enforce diversity of outputs. Our work extends to conditional synthesis and incorporates explicit diversity constraints. 

Since the purpose and definition of disentanglement in our method are different from previous work, the existing metrics for evaluating disentanglement \cite{betaVAE, metric2, metric3, metric4,style-based} are not appropriate for measuring feature disentanglement for our method. We therefore develop a new means to quantify the hierarchical disentanglement for our approach.

\section{Methods}

\noindent
Multimodal conditional image synthesis combines a given input conditional code with sampled latent codes drawn from a compact latent space (usually a standard normal or uniform distribution) and decodes the combination into an output image. Unlike previous efforts \cite{bicyclegan, muint, msgan, ndiv}, we propose a cascading disentangled decoder inspired by Laplacian image pyramid \cite{adelson1984pyramid}. With a central multi-scale backbone, it generates output images at every feature spatial scale through a single convolutional layer. We enforce the generated images at every scale to be average-pooled back into a lower-resolution generated image. By doing so, we distill features at each spatial scale to only focus on image details --- similar to a Laplacian image --- at the corresponding spatial resolution.  With these scale-independent features, we can inject random variables into each scale of the image features to model the scale-specific stochasticity of image details.



\subsection{Multiscale Disentanglement}

\noindent
To enable scale-specific editing of visual contents, we need a model that has a disentangled latent code representation. This implies that visual content on specific scales can be modified by changing corresponding latent codes, while visual contents on other scales remain unaffected. 

In a decoder network that receives only a single latent code at the coarsest scale, the single latent code controls image generation at all scales and hence changing the code will affect all scales. This motivates the multi-latent code design: injecting latent codes on all spatial scales and allowing each individual latent code to control the image on corresponding scales. Let $Z_0$ denote the random base latent code, and $Z_1 = A_1(Z_0), Z_2 = A_2(Z_0), \cdot Z_k = A_k(Z_0)$, and the latent variable at each scale level $i\in\{1...k\}$, where $A_i$ is an affine mapping matrix.  Intuitively, the latent code at coarse scale may mostly affect global structure while latent code at fine scale is more likely to alter local textureat its respective scale. Such behavior can be seen in \cite{denton2015deep} and \cite{style-based}.


However, this design does not guarantee that visual information represented by the latent code at different scales are disentangled. For example, latent code at coarse scale might control texture and color that are also affected by latent code at fine scale. With such a decoder, it is still difficult to edit the scale-specific visual information while keeping information on other scales unchanged.


Therefore, we propose a simple but effective approach to disentangle features at each layer of the decoder to only control the visual information at the corresponding spatial scale. At each layer in the decoder, we add a single convolution layer to synthesize an image at the corresponding spatial resolution. Then, we enforce that each synthesized image, when downsampled, matches the synthesized image at the previous spatial scale. We call this constraint a multi-scale disentanglement loss $L_{disent}$. Specifically, we use average pooling to downsample the synthesized image and pixel-wise Euclidean distance to constraint the downsampling consistency. Our intuition is that, by doing so, the features at each layer are not allowed to change any visual information at the previous or deeper layers. In this way, we distill each level of features to only edit the visual information at its corresponding spatial scale.

\begin{figure*}
\centering
\includegraphics[width=\textwidth]{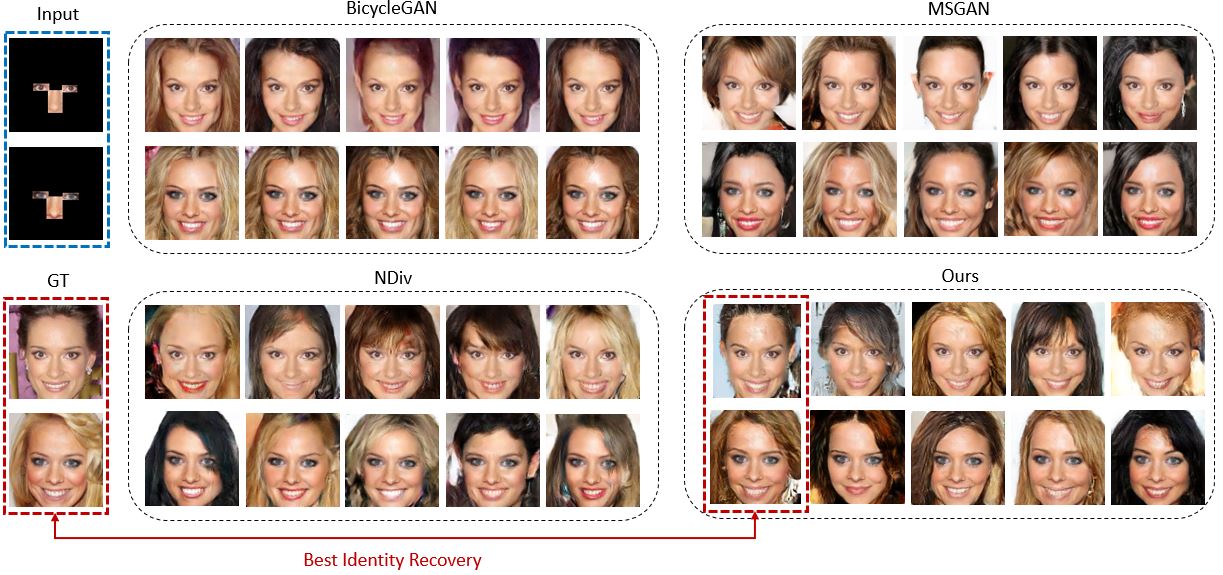}
\vspace{-20 pt}
\caption{Qualitative comparison for multimodal image outpainting. }
\label{op_fig}
\vspace{-5 pt}
\end{figure*}

\begin{table*}[t!]
\centering
\begin{tabular}{c|l|ccccc}
\toprule
\multicolumn{2}{c|}{} &  & \multicolumn{1}{r}{} & \multicolumn{3}{c}{Identity} \\ \cline{5-7} 
\multicolumn{2}{c|}{\multirow{-2}{*}{Method}} & \multirow{-2}{*}{Quality $\downarrow$} & \multicolumn{1}{r}{\multirow{-2}{*}{Diversity $\uparrow$}} & Shortest Distance $\downarrow$ & Recovery Count($\%$) $\uparrow$ & Landmark Alignment $\downarrow$ \\ \hline
\multicolumn{2}{l|}{BicycleGAN} & 64.133 & 0.093  & 0.233 & 15.34 & 5.914 \\
\multicolumn{2}{l|}{MSGAN} & 56.998 & 0.232 & 0.237 & 28.93 & 4.754 \\
\multicolumn{2}{l|}{Ndiv} & 68.855 & 0.319  & 0.256 & 20.95 & 5.126 \\
\multicolumn{2}{l|}{Ours} & \textbf{55.854} & \textbf{0.333} &\textbf{0.228} & \textbf{34.78} & \textbf{4.540} \\
\bottomrule
\end{tabular}
\vspace{-5 pt}
\caption{Quantitative Comparison with state-of-the-art approaches in multimodal image outpainting task. }
\label{op_table}
\vspace{-10 pt}
\end{table*}

Formally, we denote $S$ as the downsampling operation on image $x$, specifically, average $2\times 2$ pooling with stride of $2$. The loss function of progressive downsampling consistency is defined as follows,
\begin{equation}\label{losssub}
\begin{aligned}
  \mathcal{L}_{disent}=\sum_{i=1}^{n-1}d(S(G_{i+1}(c, z)), G_i(c, z)),
\end{aligned}
\end{equation}
where $n$ is the number of resolution scales, $c$ is the conditional code, $z$ is random variable, and $G_i$ is the generator, whose subscript refers to the network layers that are responsible for synthesizing images at each scale. For $G_i(2\leq i \leq n)$ , they have iterative format:
\begin{equation}\label{generator_format}
\begin{aligned}
  G_{i}(c,z)=U\left[ G_{i-1}(c,z),A_i(z)\right]
\end{aligned}
\end{equation}
and,
\begin{equation}\label{generator_format}
\begin{aligned}
  G_{1}(c,z)=U\left[c,A_1(z)\right].
\end{aligned}
\end{equation}
where U denotes the Upsampling Module. 

At each spatial scale, we also applied conditional GAN to synthesize photo-realistic images, where the loss functions are as follows, 
\begin{equation}\label{lossgan}
\begin{aligned}
  \mathcal{L}_{GAN}=&E_{\mathbf{x}\sim p_{\text{data}}(\mathbf{x})}\left[\sum_{i=1}^n log(D_i(S^{n-i}(\mathbf{x})|c))\right]\\
  &+E_{\mathbf{z}\sim p(\mathbf{z})}\left[\sum_{i=1}^n log(1-D_i(G_i(\mathbf{z},c)))\right].
\end{aligned}
\end{equation}

The similar multi-scale adversarial loss has been applied in SinGAN \cite{shaham2019singan} as well.



\subsection{Progressive Diversification}

\noindent
To avoid mode collapse and increase diversity of the synthesized images, we leverage the normalized diversification \cite{ndiv}, which forces the normalized pairwise distance of generative outputs to be at least as large as that of the corresponding latent variables. 


Here we introduce normalized diversification in a progressive manner, that is, we add normalized diversification loss at each layer of the hierarchical decoder. In comparison, previous work\cite{SNGAN,ndiv,msgan} applied the diversity penalty only at the final scale output of the model, which enforces in a brute force way the final output diversity but does not prevent individual levels of the model from mode collapse (for a 3-layer model where latent code $z$ is injected at each level, previous efforts only enforce that the final output varies by $z$, but allow individual layers to collapse. For example, the first layer of the model utilizes the $z$ while second and third layer ignore $z$ entirely).

Mode collapses at individual levels prevent us from exactly controlling the diversity on a specific level of structure or texture for synthesized data. Thus, we propose progressive diversification, effectively unfolding manifold topology for different scales. In this way, we achieve not only independent multi-scale control during the generative process but also guarantee latent that the code $z$ introduces variation on every scale, likely from structure to texture. 


The inserted progressive normalized diversification can be formulated as loss function \ref{Ndivloss}.

\begin{equation}\label{Ndivloss}
\begin{aligned}
  \mathcal{L}_{Ndiv}=\sum_{k=1}^n \frac{1}{N^2-N}\sum_{i=1}^N\sum_{j=1}^Nmax(\alpha D^{z}_{ij}-D^{G_k(z,c)}_{ij}),
\end{aligned}
\end{equation}

where $N$ is the number of samples to calculate the normalized pairwise distance matrix and $D^z_{ij}$, $D^{G_k(z|y)}_{ij}$ are defined as elements in normalized pairwise distance
matrix $D^z$,$D^{G_k(z|y)}\in \mathbb{R}^{N\times N}$ of ${z_i}^N_{i=1}\sim p(z)$:

\begin{equation}\label{Ndivma}
\begin{aligned}
  &D^z_{ij}=\frac{d(z_i-z_j)}{\sum_{j}d(z_i-z_j)}\\ &D^{G_k(z,c)}_{ij}=\frac{d(G_k(z,c)_i-G_k(z,c)_j)}{\sum_{j}d(G_k(z,c)_i-G_k(z,c)_j)}.
\end{aligned}
\end{equation}
Here, $d$ the latent variable is the Euclidean distance, and for generative outputs is the pixel-wise Euclidean distance.

\begin{figure*}
\centering
\includegraphics[width=\textwidth]{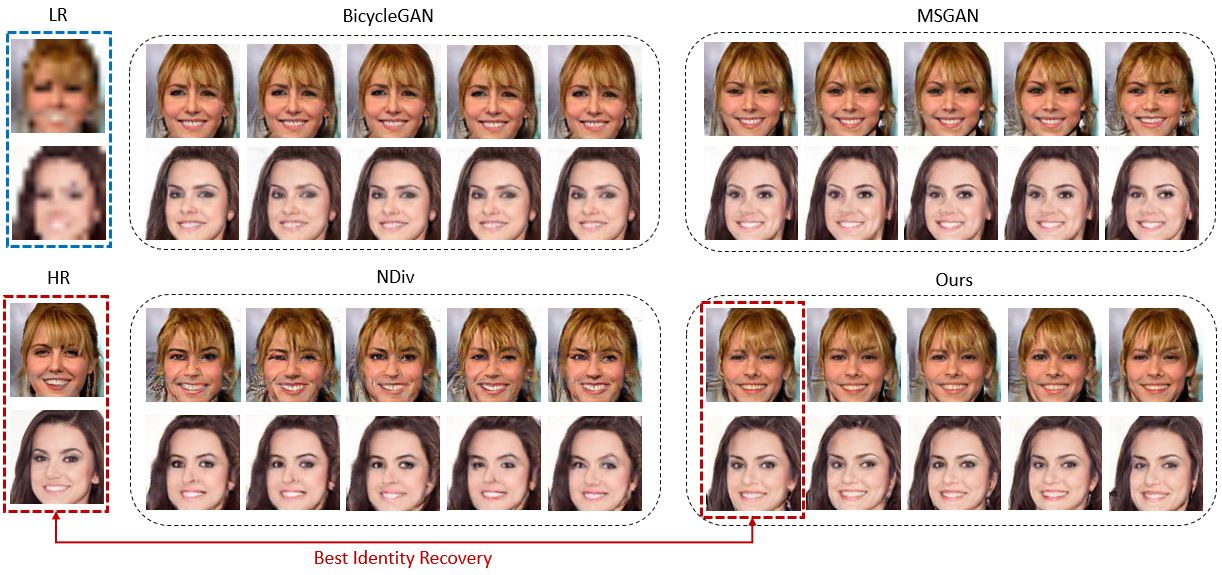}
\vspace{-20 pt}
\caption{Qualitative comparison for multimodal image superresolution}
\label{sr_results}
\vspace{-5 pt}
\end{figure*}

\begin{table*}[t!]
\centering
\begin{tabular}{c|l|ccccc}
\toprule
\multicolumn{2}{c|}{} &  & \multicolumn{1}{r}{} & \multicolumn{3}{c}{Identity} \\ \cline{5-7} 
\multicolumn{2}{c|}{\multirow{-2}{*}{Method}} & \multirow{-2}{*}{Quality $\downarrow$} & \multicolumn{1}{r}{\multirow{-2}{*}{Diversity $\uparrow$}} & Shortest Distance $\downarrow$ & Recovery Count($\%$) $\uparrow$ & Landmark Alignment $\downarrow$ \\ \hline
\multicolumn{2}{l|}{BicycleGAN} & \textbf{40.837} & 0.019  & 0.129 & 26.49 & 7.062 \\
\multicolumn{2}{l|}{MSGAN} & 49.647 & 0.0438 & 0.131 & 29.48 & 5.907 \\
\multicolumn{2}{l|}{Ndiv} & 81.4213 & \textbf{0.0758}  & 0.143 & 5.60 & 5.938 \\
\multicolumn{2}{l|}{Ours} & 46.346 & 0.0677 &\textbf{0.125} & \textbf{38.43} & \textbf{4.261} \\
\bottomrule
\end{tabular}
\vspace{-5 pt}
\caption{Quantitative Comparison with state-of-the-art approaches in multimodal image superresolution task.}
\label{sr_table}
\vspace{-10 pt}
\end{table*}

\section{Experiments}

\noindent
The proposed approaches were evaluated through their performance on various tasks, including image outpainting, image superresolution, and dog2cat/cat2dog translation. In addition to conventional quality and diversity assessment, we propose to evaluate the extent to which diverse sampling can improve identity recovery, especially in the context of facial recognition. We believe that this is a first attempt that aims to apply diverse synthesis for better recognition. Our premise is that given a conditional code containing only partial information of the ground truth image, a decoder capable of generating diverse output can produce at least one or more results that is close to or recovers the ground truth image, as long as sufficient latent code is sampled. This property would be useful in many difficult recognition situations, such as identifying criminals in largely occluded or very low-resolution images. Diverse sampling would provide a set of candidates, which can then be narrowed down further by human reviewers. We describe next three newly proposed evaluation metrics in this work for identity recovery.

\subsection{Evaluation Metrics}

\noindent
To perform evaluation of our approach, we use the following metrics.

\noindent
\textbf{FID.} We use FID \cite{FIDmetric} to evaluate the quality of generated data. This metric applies the Inception Network \cite{inceptionNN} to extract features from real and synthesized data, and then calculates the Frechet distance between the two distributions of collected real and synthesized features, respectively. A lower FID score indicates less discrepancy between real and synthesized data and hence higher quality.

\noindent
\textbf{LPIPS.} We apply LPIPS \cite{LPIPS} to quantify the diversity, which calculates the pairwise average feature distance across the whole generated dataset. We use AlexNet \cite{alexnet} pretrained on ImageNet \cite{imagenet_cvpr09} to extract features. Larger values of the pairwise LPIPS score indicate increased image diversity.

\begin{figure*}
\centering
\includegraphics[width=\textwidth]{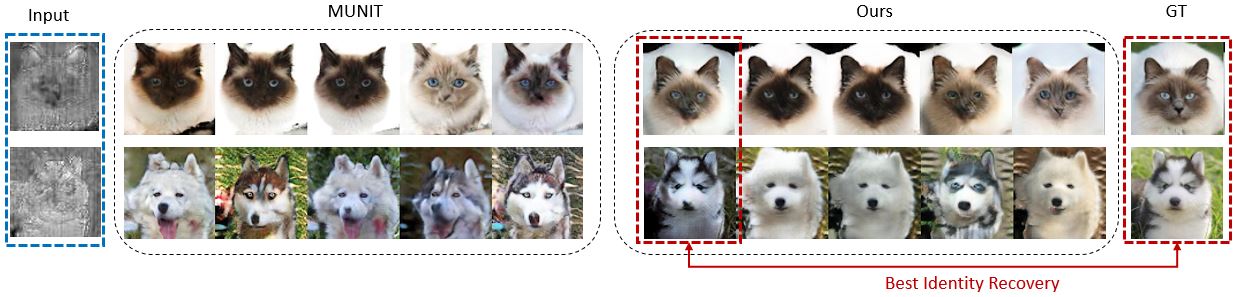}
\vspace{-10 pt}
\caption{Qualitative comparison against the state-of-the-art in multi-modal translation for cat/dog identity recovery. The input (left column) indicates a MUNIT content code. Only one channel of the content code is shown for easier . }
\label{fig:dog_results}
\vspace{-5 pt}
\end{figure*}

\begin{table*}[t!]
\centering
\begin{tabular}{l|l|cccc}
\toprule
\multicolumn{2}{c|}{} &  &  & \multicolumn{2}{c}{Identity} \\ \cline{5-6} 
\multicolumn{2}{c|}{\multirow{-2}{*}{Method}} & \multirow{-2}{*}{Quality $\downarrow$} & \multirow{-2}{*}{Diversity $\uparrow$} & Shortest Distance $\downarrow$ & Recovery Count($\%$) $\uparrow$ \\ \hline
\multicolumn{2}{l|}{MUNIT\cite{muint}} & \textbf{15.74} & 0.533  & 0.445 & 15.0 \\
\multicolumn{2}{l|}{Ours} & 21.22 & \textbf{0.547} &\textbf{0.393} & \textbf{85.0} \\
\bottomrule
\end{tabular}
\vspace{-5 pt}
\caption{Quantitative comparison with state-of-the-art approaches on the cross-modal image-to-image translation task.}
\label{tab:text_cat}
\vspace{-10 pt}
\end{table*}

\noindent
\textbf{Identity Recovery.} To quantify how well diverse sampling recovers the true facial identity, we propose to evaluate the distance between the most similar sampled output and the ground truth image. First, we compute the shortest embedding distance between a set of sampled outputs and the ground truth, where the embedding distance is given by LPIPS. We average this shortest embedding distance across all training examples, which we denote as \textbf{Shortest Distance} in Table (\ref{op_table}) and (\ref{sr_table}). We also count the chance that each method generates the most similar outputs under evaluation of this embedding distance, and denote this as \textbf{Recovery Count}. We also evaluate the identity recovery of face images using facial landmarks, which are obtained using a pretrained facial landmark detector \cite{superfan}. Specifically, the similarity between a sampled output and the ground truth is given by the mean squared error distance between the corresponding 68 facial landmark locations for the sampled and ground-truth images. We refer to this evaluation as \textbf{Landmark Alignment}. 



\subsection{Image Outpainting}

\noindent
We first present experimental results in an image outpainting task, where the goal is to fill in large areas of missing pixels in a highly occluded image, which may have many different solutions for a given input. In terms of model implementations, we only need to add our proposed multi-scale disentanglement and diversification into a standard conditional encoder-decoder. The experiment is conducted over the CelebA dataset \cite{celeba} with cropped 128x128 images. 

We compare our model with the current state-of-the-art multimodal conditional sysnthesis approaches, including BicycleGAN \cite{bicyclegan}, MSGAN \cite{msgan}, and NDiv \cite{ndiv}. The experimental results show that our model can generate the best results in terms of image quality, diversity as well as identity, as shown in Table.\ref{op_table}. In our qualitative comparison figure.\ref{op_fig}, we show that one of the sampled image could best recover the ground truth facial identity. We think that this is because our model can sample not only very diverse but also realistic images. 

\begin{figure*}
\centering
\includegraphics[width=0.9\textwidth]{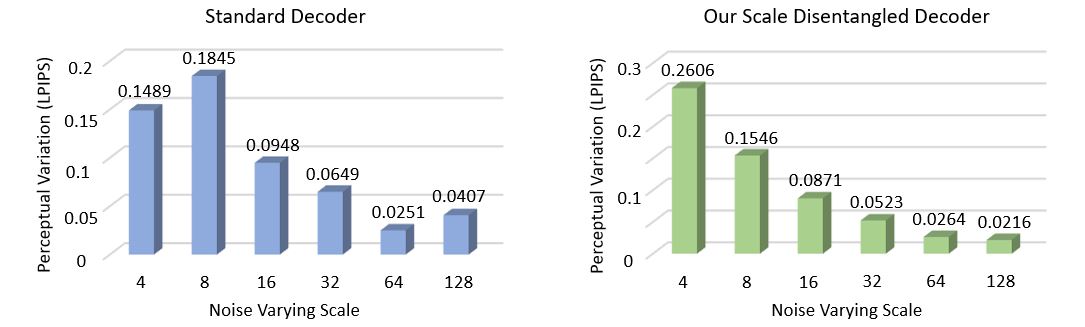}
\caption{This figure demonstrates how much the generated image would change by varying noise injection at a specific resolution scale. This is implemented by sampling a reference code, and only varying random variable at one scale at a time and fixing the random variables at the other scales constant. }
\label{fig:main}
\vspace{-10 pt}
\end{figure*}

\subsection{Image Super-Resolution}

\noindent
Another multimodal conditional synthesis we run on face data is super-resolution. While most of other super-resolution approaches model this task as a deterministic image-to-image process, we consider super-resolution from a very low resolution image as a one-to-many process because of its uncertainty in nature. For example, the 16x16 low resolution in Figure.\ref{sr_results} can be very difficult to identified. Even for human, it is very difficult to tell what is the ground true high-resolution image is. Or, every human would probably have a different answer. In this task, we use bilinearly downsampled 16x16 image as low-resolution input and 128x128 image as high-resolution output in CelebA dataset \cite{celeba}.

Our implementation of superresolution model is similar to image outpainting, but we start to decode image at scale of 16x16, which is the same as input resolution. The disentanglement and diversification is added at every other higher resolution scale. As seen from Table. (\ref{sr_table}), even though our approach does not reach the best quality or diversity, but it achieves the best identity score. We think that the simply measuring quality or diversity separately is not sufficient. BicycleGAN has the best quality but lack diversity, so it is difficult to “hit” the ground truth image. NDiv has the largest diversity but lacks quality, and thus it is also very difficult to recover the realistic ground truth image. In contrast, our model can produce reasonably large diversity and good quality, and thus has the highest chance to recover the ground truth image. Our implementation of super-resolution model is similar to image outpainting, but we start to decode image at scale of 16x16, which is the same as input resolution. The disentanglement and diversification is added at every other higher resolution scale. As seen from Table. (\ref{sr_table}), even though our approach does not reach the best quality or diversity, but it achieves the best identity score. We think that the simply measuring quality or diversity separately is not sufficient. BicycleGAN \cite{bicyclegan} has the best quality but lack diversity, so it is difficult to “hit” the ground truth image. NDiv \cite{ndiv} has the largest diversity but lacks quality, and thus it is also very difficult to recover the realistic ground truth image. In contrast, our model can produce reasonably large diversity and good quality, and thus has the highest chance to recover the ground truth image.



\subsection{Dog to Cat Image Translation}

\noindent
We conduct an unpaired cross-modal image translation using the MUNIT backbone\cite{muint} on the cat and dog dataset from \cite{drit}. This dataset contains 1264 training/100 testing dog and 771 training/100 testing cat images. We modify the MUNIT backbone for multi latent codes injection and implemented the disentanglement loss and progressive diversification loss as mentioned in section 3.1 and 3.2. Detailed network architecture before and after modification can be found in the supplemental material. We compare FID and LPIPS of our model against the original MUNIT model trained on the cat and dog dataset and report the performance in Table.\ref{tab:text_cat}. For identity recovery evaluation, we pass the ground truth image through the content encoder from the MUNIT framework and derive a content code, which is recombined with sampled latent codes and decoded into images. Quantitatively, we evaluate identity recovery using the LPIPS metric from 4.1. Qualitative results of identity recovery is demonstrated in Fig.\ref{fig:dog_results}.

\subsection{Multi-scale Disentanglement Evaluation}

\noindent
To quantify the multiscale disentanglement, we evaluate perceptual variations on output images against different noise varying scales. In specific, for scale $k$ in our $n$ scale decoder, a given fixed input condition code $c$, we sample a center latent code $z$. We fixed the input latent codes on every scale to be $z$ except for for scale $k$, where we sample 10 latent codes centered around $z$ and generate 10 different output images. Pairwise perceptual distances (using LPIPS) among these 10 images are calculated. We calculate the pairwise perceptual distance for scale $k$ across 1000 different input condition codes and averaged them. We plotted the averaged perceptual distance against the scale on Fig.\ref{fig:main}. For a general decoder (left) without the the disentanglement constraint, the perceptual variation were not monotonically decreasing along the scale. Our scale disentangled decoder (right) on the other hand achieves multiscale disentanglement: latent codes at finer scales monotonically introduce less variation to the image, therefore editing latent code at finer level has little effect on visual information on coarse scale.

\section{Conclusion}

\noindent
We develop a conditional image synthesis network that enables scale-specific and diverse control of image content. We instantiate our design with a cascading decoder network. We couple it with multi-scale feature disentanglement constraints and a progressive diversification regularization. In addition, we gain semantic persistency in the decoder by sharing latent code across scales during training. This allows for stratified navigation and search within latent code space, and motivate the task of identity recovery. We propose three evaluation metrics for identity recovery within conditional image synthesis scenarios. On tasks of image outpainting, image superresolution, and multimodel image translation, our method consistently outperforms state-of-the-art counterparts in terms of identity recovery, while having competitive image quality and diversity. Hence we believe our method may potentially be useful in extreme image recognition situations, such as recognizing criminals in largely occluded or very low-resolution images, and finding lost pets from low quality surveillance images. 

\section*{Acknowledgement}
\vspace{-5 pt}

We gratefully appreciate the support from Honda Research Institute Curious Minded Machine Program. We also gratefully acknowledge a GPU donation from NVIDIA. 


{\small
\bibliographystyle{ieee_fullname}
\bibliography{egbib}
}











\twocolumn[\fontsize{14}{14}\textbf{Supplementary Materials: Nested Scale-Editing for Conditional Image Synthesis}
\newline
]


\appendix

In this section, we discuss the implementation details and demonstrate more qualitative results of our experiments on multimodal image outpainting, image super-resolution, cross-domain image translation, and text-to-image translation.

\section{Image Outpainting}

The detailed implementation of our decoder network architecture for image inpainting is shown in Fig.3 in the main paper. At each spatial scale, the number of channels for feature activation is 512. The conditional code is the feature vector of the occluded image encoded by a standard encoder network. The detailed implementations of the encoder network are listed in Table \ref{encoder}. We use negative slope of 0.2 for all LeakyReLU layers throughout the network. We employ the following abbreviation: N = Number of filters, K = Kernel size, S = Stride, P = Padding. "Conv" and "SN" denote convolutional layer and instance normalization respectively.

\begin{table}[!h]
\centering
\begin{tabular}{lll}
\toprule
 \textbf{Layer} & \textbf{Hyper-parameters}  \\
\midrule
 1  & Conv(N64-K4-S2-P1) + LeakyReLU \\
 2  & Conv(N128-K4-S2-P1) + IN + LeakyReLU \\
 3  & Conv(N256-K4-S2-P1) + IN + LeakyReLU \\
 4  & Conv(N512-K4-S2-P1) + IN + LeakyReLU \\
 5  & Conv(N256-K4-S2-P1) + IN + LeakyReLU \\
 6  & Conv(N256-K4-S2-P1) + IN + LeakyReLU \\
 7  & Conv(N128-K1-S2-P1) + LeakyReLU \\ 
\bottomrule
\end{tabular}
\caption{Encoder network for image outpainting and super-resolution. }
\label{encoder}
\end{table}

The weights for the adversarial loss, disentangle loss, and diversity loss are all set to be ones.  To enforce the diversity of synthesis, we sample $N=4$ random variables at each iteration. We set the relaxation hyperparemeter $\alpha$ in the diversity hinge loss to be 0.8. With batch size of 24, we train the network using Adam optimizer \cite{kingma2014adam} with learning rate of 2e-4, beta1 of 0.5, beta2 of 0.999.

\section{Image Super-Resolution}

Our super-resolution network is mostly similar to the network used for image outpainting with two major differences. The first difference is that we do not decode any image lower than the low-resolution scale (16x16), since there is no need to edit visual details below the input resolution scales. Thus, our decoder starts to generate images at scale of 32x32 and enforces the downsampled sample of the 32x32 images to be the same as the ground-truth 16x16 low-resolution image. The disentanglement loss for scales of 64 and 128 are the same as the outpainting newtork. In addition, we add skip connections from the encoder to the decoder for the purpose of preserving low-resolution structural information. The encoder for the low-resolution image is the same as the encoder used in image outpainting, which is shown in Table \ref{encoder}. We also use the same optimizer and hyperparameters for both the image super-resolution and image outpainting.

\section{Cross-Domain Translation}

For the cross-domain translation task we adapted the MUNIT network\cite{muint}. In terms of network architecture, we use exactly the same content and style encoders as the original and we only modify the decoder, where we add an additional convolution for image output at the 128x128 resolution, and correspondingly the discriminator for it. We used the default multi-resolution discriminator as in the original implementation. For details of the architecture we refer reader to \cite{muint} and its official github repository \footnote{https://github.com/NVlabs/MUNIT.}. In terms of losses, in addition to the original reconstruction losses and discriminator losses, we calculated the proposed disentangle loss $L_{disent}$ between the two levels as well as the normalized diversity loss \cite{ndiv} on each level. We use the following weights for losses: weight of adversarial loss $\lambda_{GAN} = 1$; weight of image reconstruction loss $\lambda_{xw} = 10$; weight of style reconstruction loss $\lambda_{sw} =1$; weight of image reconstruction loss $\lambda_{cw} = 1$; weight of normalized diversity loss $\lambda_{ndiv} =1$; weight of the disentangle loss $\lambda_{disent} =1$. An illustration of the network architecture as well as the added losses is shown in Fig.\ref{fig:munit_arch}. We optimize the network using an Adam optimizer with learning rate of $1e-4$, $beta1$ of 0.5 and $beta2$ of 0.999 with batch size of 2.

\section{Text-to-Image synthesis}

Other than the image outpainting, image superresolution and cross-domain translation, we also evaluate our proposed multi-scale disentangle loss and the normalized diversity loss on the task of text-to-image synthesis. Our implementation is based on the StackGAN++ \cite{zhang2018stackgan++}. We refer interested reader to \footnote{https://github.com/hanzhanggit/StackGAN-v2} for the original implementation. We use pretrain text embedding from \cite{reed2016generative}, as in \cite{zhang2018stackgan++} and \cite{msgan}. We keep the original text embedding sampling unchanged but incorporate two changes within the decoder. First, we incorporate the adaIn layer \cite{huang2017arbitrary} for each refine stage, which allows injection of random latent vector at each stage. In comparison the original implementation only inject latent random vector at the init stage. Second, we added two more level of image output to the original image. The original StackGAN network outputs 64/128/256 images. We extended the network to output 16/32 images. With the two changes in place, we add the proposed disentangle loss and normalized diversification loss to it. Detailed architecture of the modified StackGAN++ is illustrated in Fig.\ref{fig:stackgan_arch}. We tested our network on the cub\_200\_2011 \cite{WahCUB_200_2011} birds dataset. As in the image outpainting, image superresolution and cross-domain translation task, we achieve scale-specific editing by injecting different latent codes at each scale at test time, as shown on \ref{fig:text2im}. Quantitatively, our network achieved similar image quality (measured by FID) and slightly higher diversity (measured by LPIPS) as previous state-of-the-art from \cite{msgan}, as shown by Table.\ref{tab:stackgan}. We optimize the network using an Adam optimizer with learning rate of $2e-4$, $beta1$ of 0.5 and $beta2$ of 0.999 with a batch size of 4.

\begin{table}[t!]
\centering
\begin{tabular}{l|cc}
\toprule
Method & {Quality $\downarrow$} & {Diversity $\uparrow$}  \\ \hline
{MSGAN\cite{msgan}} & \textbf{18.64} & 0.661 \\
{Ours} & 20.88 & \textbf{0.668} \\
\bottomrule
\end{tabular}
\caption{Quantitative comparison with state-of-the-art approaches on the cross-modal image-to-image translation task.}
\label{tab:stackgan}
\end{table}


\begin{figure*}
\centering
\includegraphics[width=0.7\textwidth]{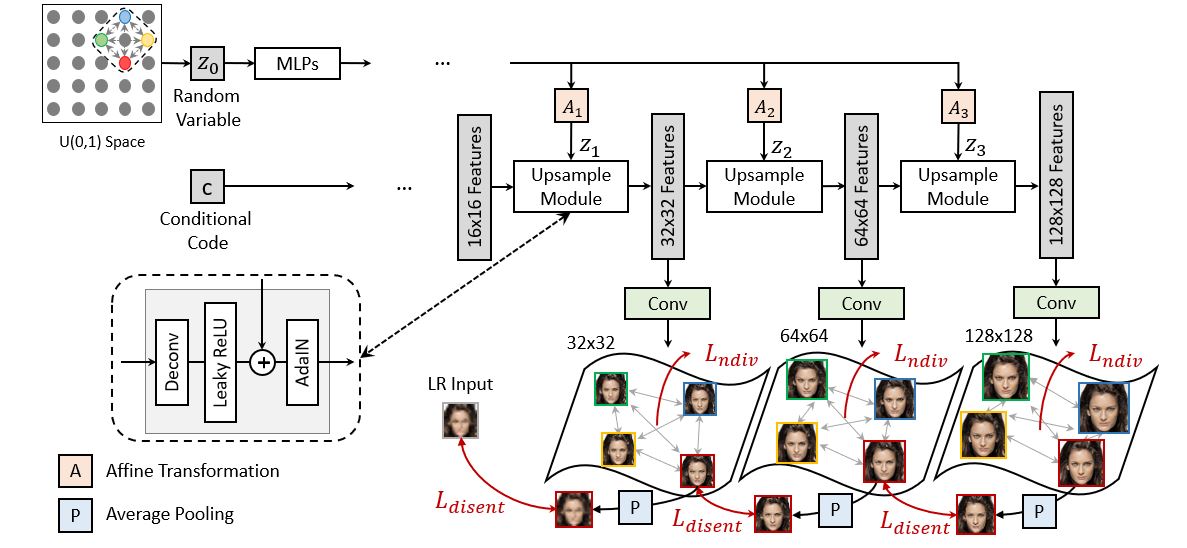}
\caption{The model architecture of super-resolution decoder network. }
\label{fig:main}
\end{figure*}
 
\begin{figure*}
\centering
\includegraphics[width=0.7\textwidth]{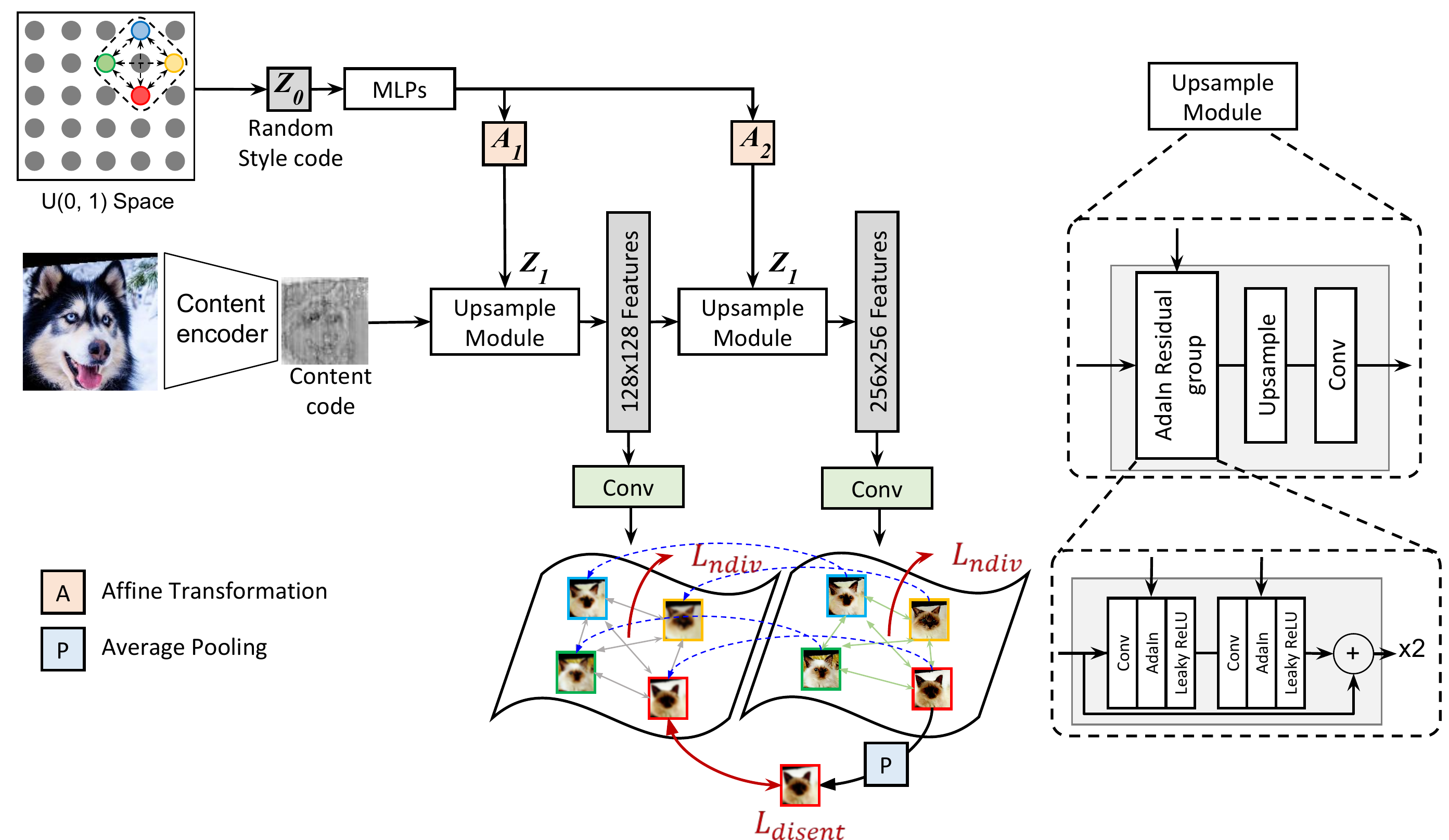}
\caption{The model architecture of modified MUNIT \cite{muint} decoder network.}
\label{fig:munit_arch}
\end{figure*}
 
\begin{figure*}
\centering
\includegraphics[width=0.7\textwidth]{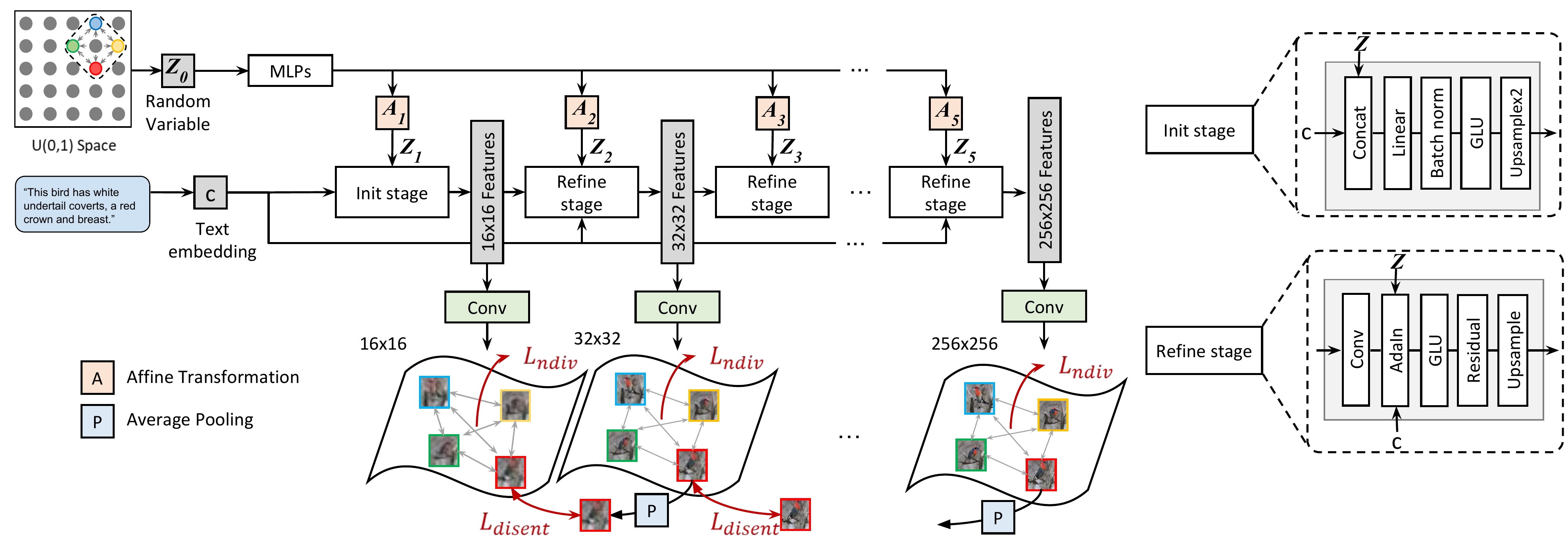}
\caption{The model architecture of modified StackGAN++ \cite{zhang2018stackgan++} decoder network.}
\label{fig:stackgan_arch}
\end{figure*}

\begin{figure*}
\centering
\includegraphics[width=0.7\textwidth]{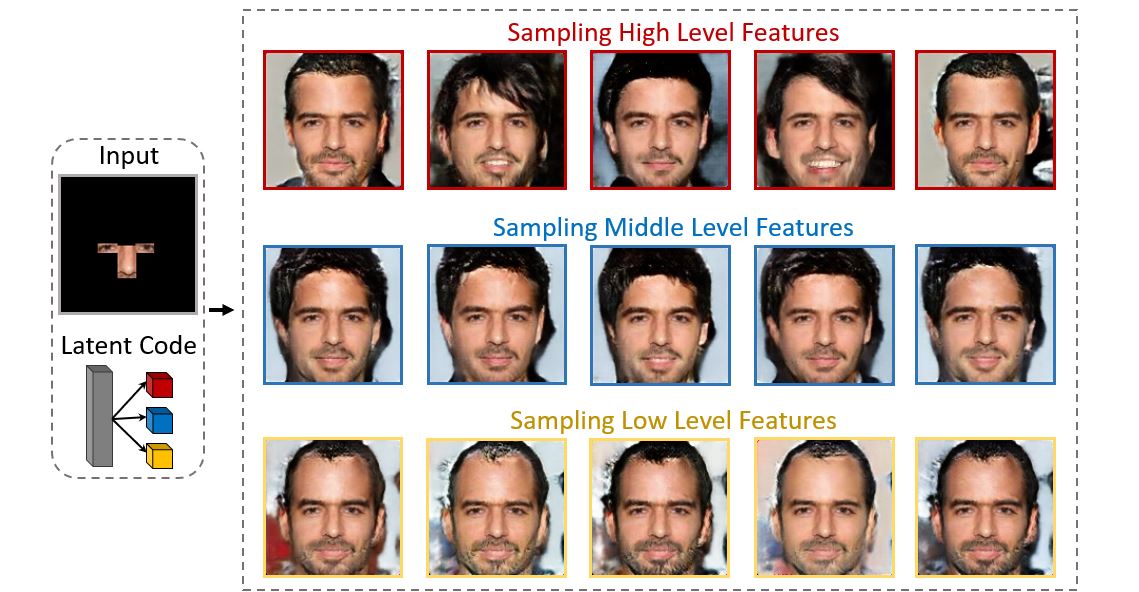}
\includegraphics[width=0.7\textwidth]{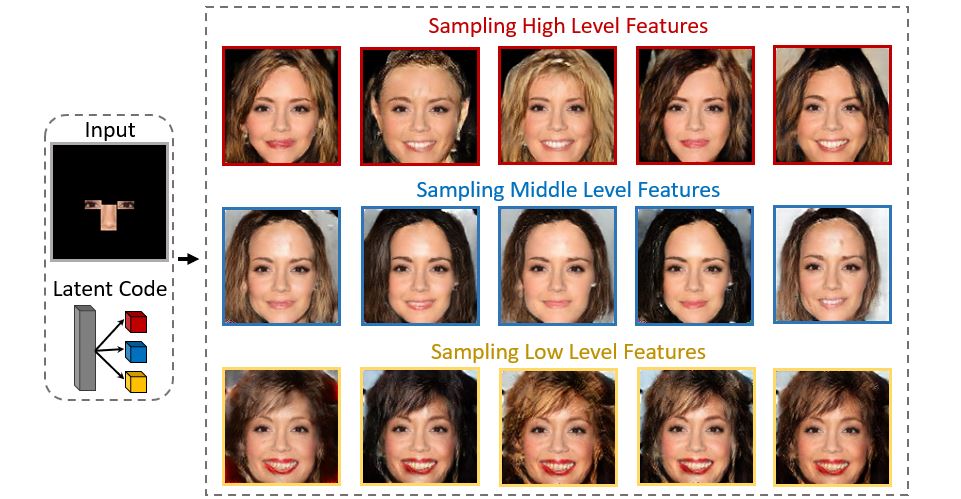}
\includegraphics[width=0.7\textwidth]{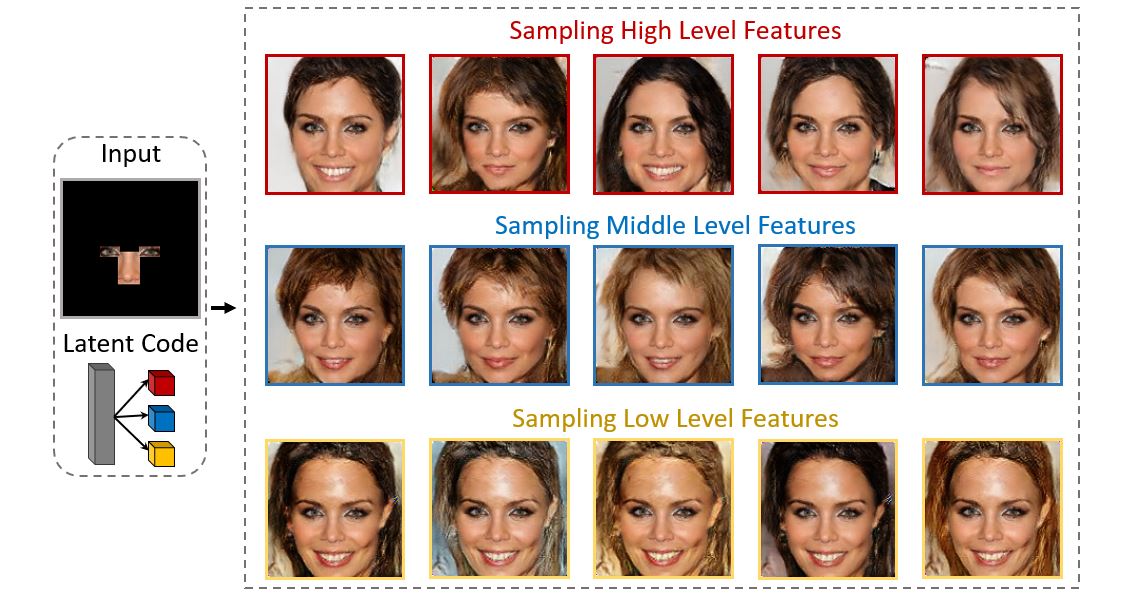}
\caption{Qualitative results for scale-editing for image outpainting. We vary random variables at scale of 4 to edit high-level features, vary random variables at scale of 8 and 16 to edit the middle-level features, and vary random variables at scales of 32, 64, 128 to edit the low-level features.  }
\label{more_detection}
\end{figure*}

\begin{figure*}
\centering
\includegraphics[width=0.7\textwidth]{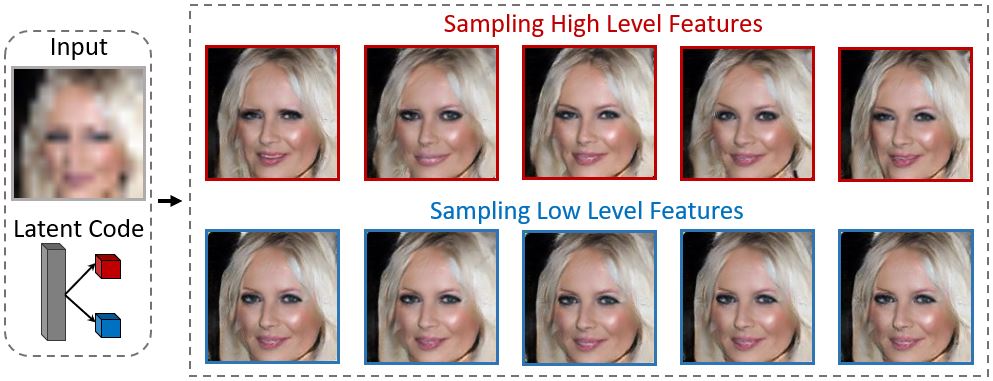}
\includegraphics[width=0.7\textwidth]{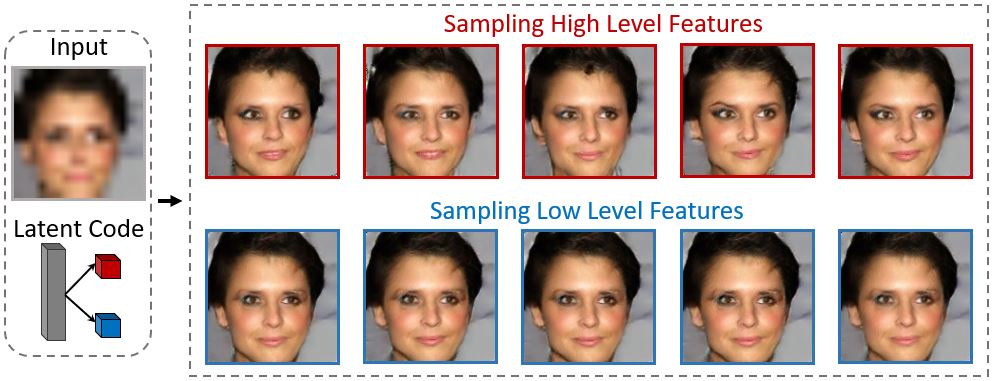}
\includegraphics[width=0.7\textwidth]{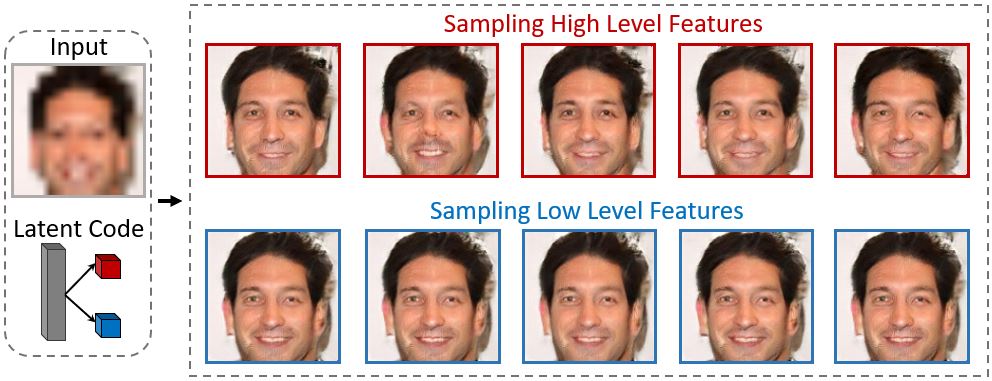}
\includegraphics[width=0.7\textwidth]{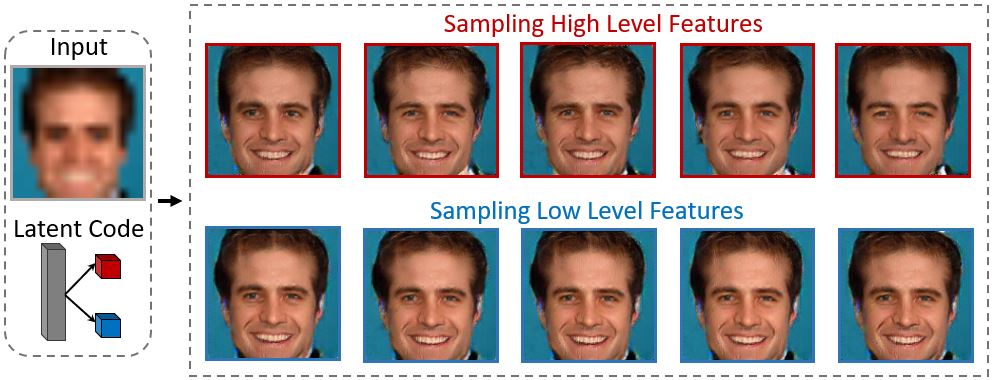}
\caption{Qualitative results for scale-editing for image super-resolution. We vary random variables at scale of 32 to edit high-level features, and vary random variables at scales of 64 and 128 to edit the low-level features. Note that the variations for this task are small in nature, and the low-level features in this super-resolution task only affect subtle textures. In the super-resolution task, our main goal is to generate multimodal outputs while preserving identities.  }
\label{more_detection}
\end{figure*}

\begin{figure*}
\centering
\includegraphics[width=0.7\textwidth]{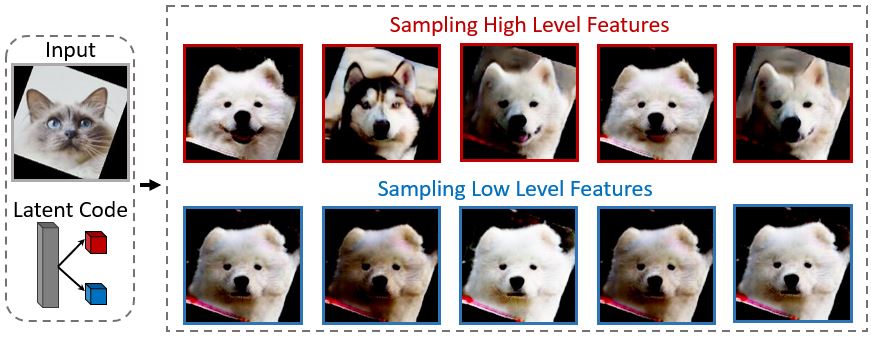}
\includegraphics[width=0.7\textwidth]{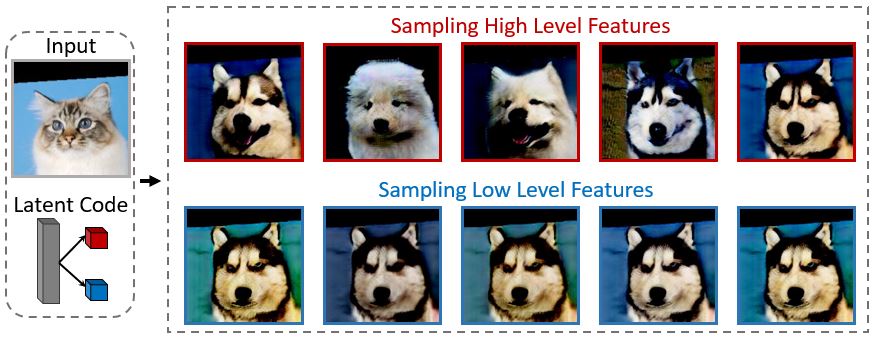}
\includegraphics[width=0.7\textwidth]{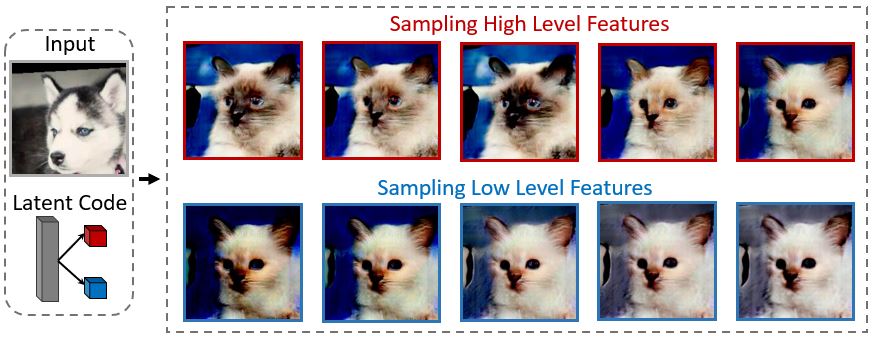}
\includegraphics[width=0.7\textwidth]{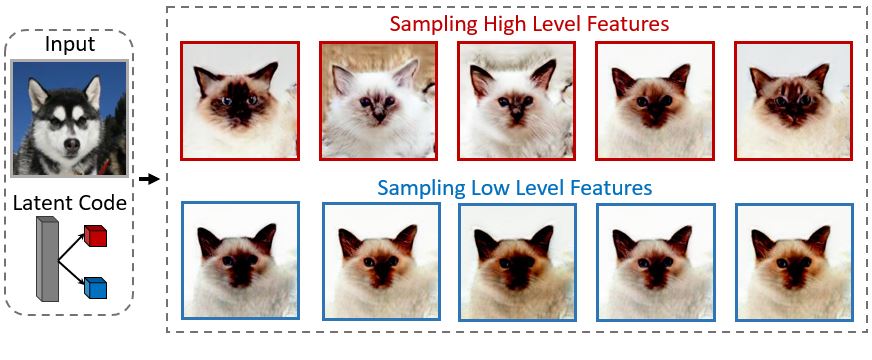}
\caption{Qualitative results for scale-editing for cat2dog and dog2cat translation. Note that there are few modes of variation compared to other tasks, due to 1). primarily a small dataset size (871 cat and 1364 dog images) and 2). there are only two types of dogs (husky and samoyed) and mostly one type of cat (siamese) in the dataset. }
\label{more_detection}
\end{figure*}

\begin{figure*}
\centering
\includegraphics[width=0.7\textwidth]{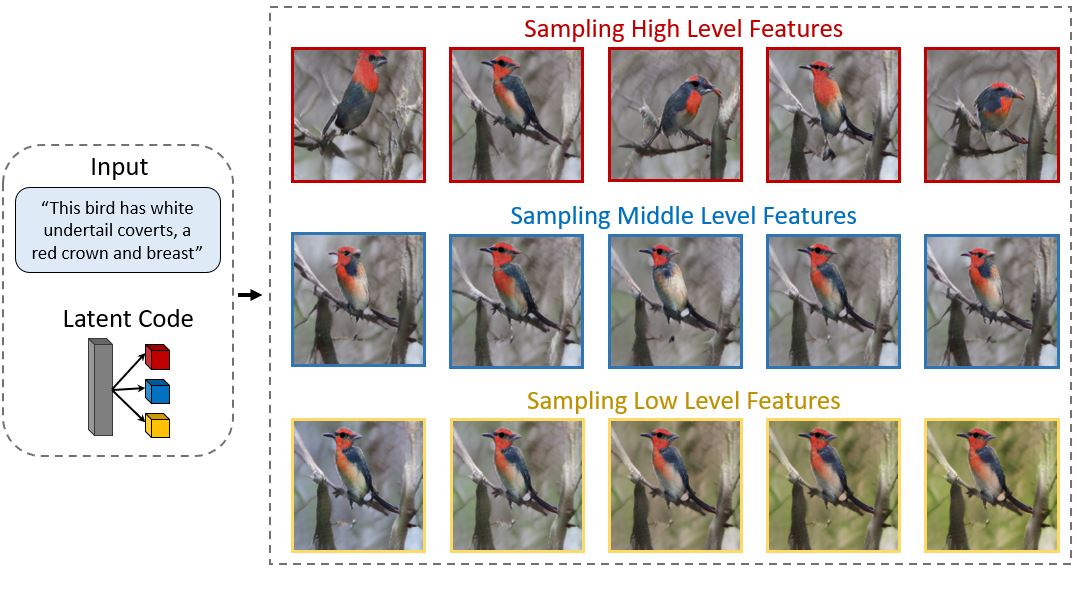}
\includegraphics[width=0.7\textwidth]{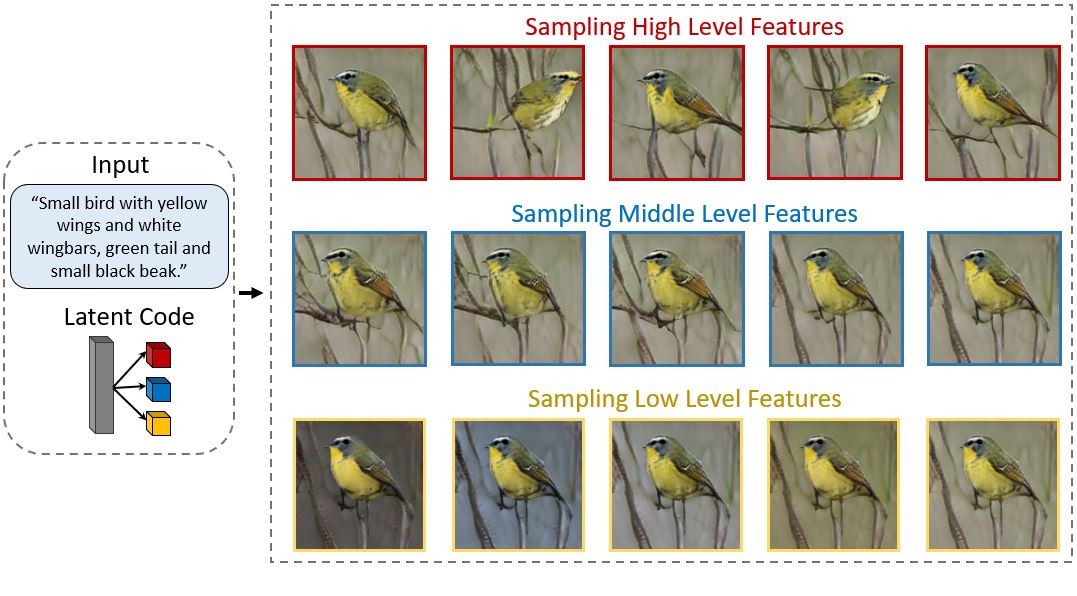}
\includegraphics[width=0.7\textwidth]{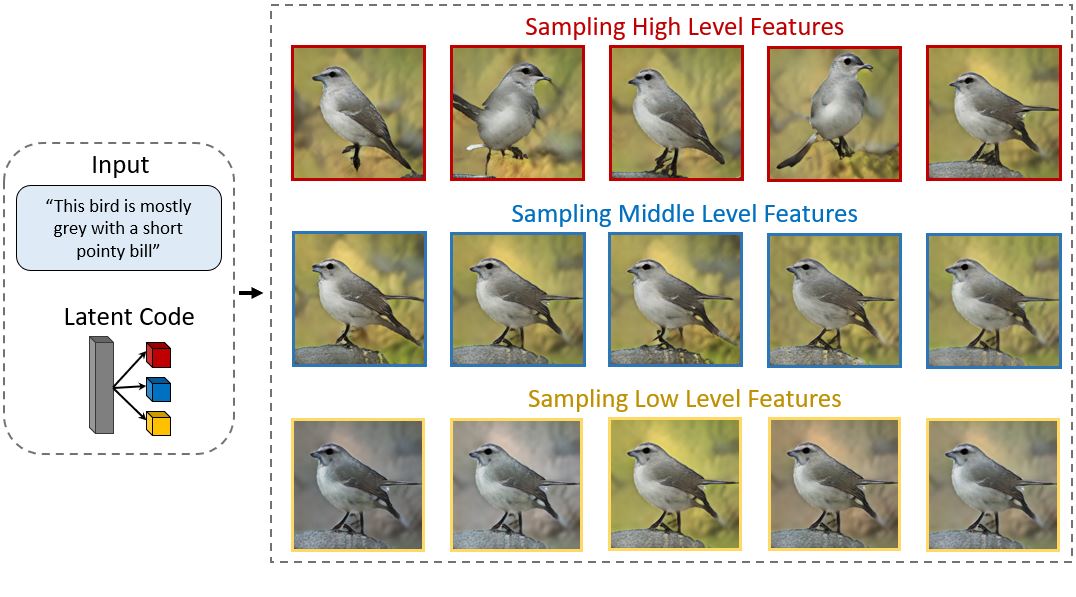}
\caption{Qualitative results for scale-editing for text-to-image translation. }
\label{fig:text2im}
\end{figure*}





\end{document}